# Vision-based Pedestrian's Potential Risk Analysis based on Automated Behavior Feature Extraction for Smart and Safe City

**Byeongjoon Noh[1] (powernoh@kaist.ac.kr), Dongho Ka[1] (kdh910121@kaist.ac.kr), David Lee[1] (david733@kaist.ac.kr), and Hwasoo Yeo[1]\* (hwasoo@kaist.ac.kr)**

[1]: Department of Civil and Environmental Engineering, Korea Advanced Institute of Science and Technology, 291 Daehak-ro, Yuseung-gu, Daejeon, South Korea

\**corresponding author*

## Abstract

Despite recent advances in vehicle safety technologies, road traffic accidents still pose a severe threat to human lives and have become a leading cause of premature deaths. In particular, crosswalks present a major threat to pedestrians, but we lack dense behavioral data to investigate the risks they face. Therefore, we propose a comprehensive analytical model for pedestrian's potential risk using video footage gathered by road security cameras deployed at such crossings. The proposed system automatically detects vehicles/pedestrians, calculates trajectories by frames, and extracts behavioral features affecting the likelihood of potentially dangerous scenes between these objects. Finally, we design a data cube model by using the large amount of the extracted features accumulated in a data warehouse to perform multi-dimensional analysis for potential risk scenes with levels of abstraction, but this is beyond the scope of this paper, and will be detailed in a future study. In our experiment, we focused on extracting the various behavioral features from multiple crosswalks, and visualizing and interpreting their behaviors and relationships among them by camera location to show how they may or may not contribute to potential risk. We validated feasibility and applicability by applying it in multiple crosswalks in Osan city, Korea.

# 1. Introduction

Despite advances in vehicle safety technologies, road traffic accidents globally still pose a severe threat to human lives and have become a leading cause of premature deaths [1]. Every year, approximately 1.2 million people are killed and 50 million injured in traffic accidents [2], [3]. Pedestrians are especially exposed to various hazards, like drivers failing to yield to them in crosswalks [2]. According to international institutes such as British Transport and Road Research Laboratory and World Health Organization (WHO), crossing roads at unsignalized crosswalks is as dangerous for pedestrians as crossing roads without crosswalks or traffic signals [4].

There are various ways to protect road users from traffic accidents, such as suppressing dangerous or illegal behaviors of road users (drivers and pedestrians) by deploying speed cameras and fences, and operating 24-hour CCTV surveillance centers at administrative districts. In addition, studies have analyzed actual traffic accidents and their factors [5], [6]. These used historical accident data or metadata to improve the safety of road environments post-facto. Alternatively, we can analyze some factors for potential risks before traffic accidents occur, in order to prevent accidents proactively by deploying speed cameras, speed bumps, and other traffic calming measures.

A breakthrough for proactive traffic safety is to analyze pedestrian's potential risks (e.g. near-miss collision). To date, an extensive variety of studies have reported on these systems [2], [7]–[11] based on vision sensors, such as closed-circuit televisions (CCTVs), which have already been deployed on roads for multiple purposes. The use of vision sensors is supposed to make it easier to study potential traffic risks over long periods of time, and allows analyses such as evaluating the behavioral factors that pose a threat to pedestrians at crosswalks based on vehicle-pedestrian interactions [7], [8], [12], [13], and supporting decisions based on their subtle interactions [9], [10].

In general, one of the most important steps in vision-based traffic safety and surveillance systems is to obtain the behavioral features of vehicles and pedestrians from the video footage. However, since most CCTVs are already deployed with oblique views of the road, it is difficult to obtain precise coordinates and behavioral features such as objects' speeds and positions. Thus, many studies used manual inspection to reliably extract these features from video footage. This requires more cost and time when extended to the urban scale, so we should address these challenges when seeking to analyze pedestrian safety across many sites in the city.

In this study, we propose a comprehensive analytical model for pedestrian's potential risk scenes. The proposed system uses data collected from existing CCTV cameras, which are normally installed for non-traffic-safety purposes such as street crime monitoring and congestion measurement. The objectives of this study are: (1) to extract the video clips with motion from the entire footage using frame difference method; (2) to automatically extract objects' trajectories over frames, and their behavioral features affecting the likelihood of potential dangerous scenes between vehicles and pedestrians; and (3) to analyze behavioral features and relationships among them by camera location. This study follows earlier experiments using sample footage [14], but improves on the methods and expands to a much larger dataset covering nine cameras over two weeks. The rest of this paper is structured as follows:

1. Materials and methods: Overview of our video dataset, methods of preprocessing images into trajectories, and extracting objects' behavioral features.

2. Experiments and results: Validation of preprocessing results, analyzing the objects' behavioral

features by spots, and discussion of results and limitations.

3. Conclusion: Summary of our study and future research directions.

This study has these novel contributions: (1) Recasting CCTV cameras for surveillance to contribute to the study of pedestrian environments with a large dataset of vehicle-pedestrian interactions; (2) performing time-based analyses of the extracted features by spots; and (3) creating one sequential process from detecting objects to extracting their behavioral features, and further analyzing them. We validate the feasibility of this model by applying it to video footage collected from crosswalks in various conditions in Osan city, Korea. This research is the first step in our long-term effort to develop a comprehensive analytical model for pedestrian's potential risk scenes that can help us understand these scenes as spatial-temporal patterns, and help identify unsafe pedestrian environments in cities.

## 2. Materials and Methods

The proposed system consists of four modules: 1) data sources; 2) preprocessing; 3) objects' behavioral feature extraction; and 4) multidimensional analysis, as seen in Figure 1. In the first module, the motion-scenes are extracted from the video streams using frame difference methods, then we detect the traffic-related objects from each frame using the mask R-CNN (Regional Convolutional Neural Network) model.

Next, we recognize the "ground tip" points of vehicles and pedestrians, which are situated directly underneath the front bumper and on the ground between the feet, respectively. We transform these ground tip points into the overhead viewpoints, and then extract various behavioral features at the scene-level, such as vehicle speed, vehicle acceleration, pedestrian safety margins (PSMs), and pedestrian position changes. In order to obtain these behavioral features, it is essential to identify each object in consecutive frames by tracking it. Thus, we implement a tracking and indexing algorithm using distance-threshold and closest-distance methods based on Kalman filter.

In the last module, the large amount of the extracted features is accumulated in a data warehouse, and we construct a data cube model based on the fact tables. Then, we perform multi-dimensional analysis for potential risk scenes with levels of abstraction. This module, and the design of the user interface for exploring and operationalizing the data, is beyond the scope of this paper, and will be detailed in a future study.

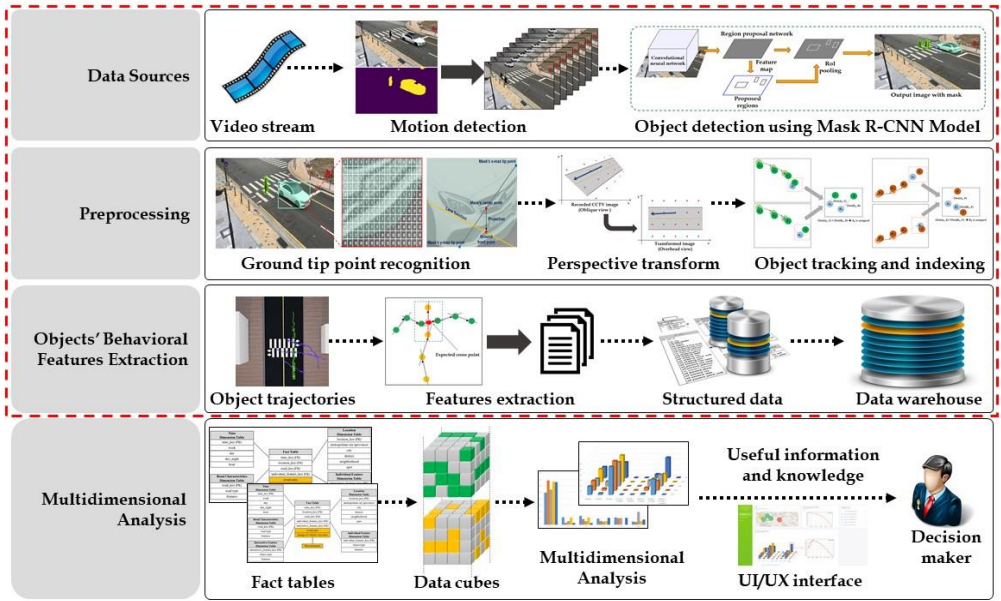

**Figure 1.** Overall architecture of the proposed system and the parts focused on in this paper (red box)

## 2.1. Data Sources

In our experiments, we used video data from CCTV cameras deployed on nine roads in Osan city, Korea. The information for each spot is arranged in Table 1, including road characteristics and recording metadata. These cameras are deployed over crosswalks at unsignalized crosswalks, and are intended to record and deter instances of street crime. Some are deployed in school zones, which are certain roads near facilities for children under age 13, e.g. elementary schools, daycare centers and tutoring academies. Penalties for breaking traffic rules or causing accidents in these areas are highly severe, such as fines of up to 3,000 million won or life imprisonment, in order to suppress risky behavior [15].

All video frames were processed locally on a computer server we deployed in the Osan Smart City Integrated Operations Center, and we only obtained the processed trajectory data, after removing the original video data. This was to protect the privacy of anyone appearing in the footage. Future systems could employ internet-connected cameras that process images on-device in real-time, and transmit only trajectory information back to servers.

Figure 2 (a) to (i) show the CCTV views being actually recorded in spots A to I, respectively. Since these spots have high "floating population" during commuting hours, due to their proximity to schools and residential complexes, we used video recorded on weekdays from January 9th to January 28th 2020, from 8 am to 10 am and from 6 pm to 8 pm.

Next, we extracted only video clips with moving vehicle activity, regarded as "scenes". In general, video streams had long idle periods, and motioned-scenes occurred occasionally as described in Figure 3. We required a simple and low computational complexity method to handle the huge volume of video data recorded in multiple spots during long time periods (approximately 504 hours, or 9 spots × 14 days × 4 hours per day).

**Table 1.** Information of the obtained spots

| Spot code | Cam. name | Crosswalk length (m) | School zone | Speed cam. | # of lanes | Signal light | Speed limit (km/h) | Frame size | Frame-per-sec (FPS) |
|---|---|---|---|---|---|---|---|---|---|
| A | Unam Elemantary school, back gate #2 | about 8 m | ✓ | × | 2 | × | 30 km/h | 1920×1080 | 25 |
| B | Yangsan Elementary school, main gate #1 | about 11 m | ✓ | × | 3 | × | 30 km/h | 1920×1080 | 25 |
| C | Gohyeon Elementary school, back gate #2 | about 20 m | ✓ | × | 4 | × | 30 km/h | 1920×1080 | 25 |
| D | Municipal Southern Welfater/Daycare center #3 | about 7 m | ✓ | × | 2 | ✓ | 30 km/h | 1280×720 | 30 |
| E | iFun daycare center #2 | about 8 m | ✓ | × | 2 | ✓ | 30 km/h | 1280×720 | 30 |
| F | Daeho Elementary school opposite side #3 | about 23 m | ✓ | ✓ | 4 | × | 30 km/h | 1280×720 | 30 |
| G | Segyo complex #9 back gate #2 | about 8 m | × | × | 2 | ✓ | 30 km/h | 1280×720 | 15 |
| H | iNoritor daycare center #2 | about 8 m | ✓ | × | 2 | ✓ | 30 km/h | 1280×720 | 11 |
| I | Kids-mom daycare center #3 | about 7 m | ✓ | × | 2 | ✓ | 30 km/h | 1920×1080 | 25 |

**Note**.: ✓: Yes  ×: No

Therefore, we applied a frame difference method, a widely used approach for detecting the moving objects from the fixed cameras [16], [17]. This method simply calculates the pixel-based difference between two frames, as an image obtained at the time t, denoted by I(t) and the background image denoted by B:

$$P[F(t)] = P[I(t)] - P[B] \qquad (1)$$

where pixel value in I(t) denoted by P[I(t)], and P[B] means the corresponding pixels at the same position on the background frame. As a result of frame difference, we can observe an intensity for the pixel positions which have changed in the two frames, and then detect the "motion" by comparing it with threshold as follows:

$$|P[I(t)] - P[I(t+1)]| > Threshold \qquad (2)$$

If motion is detected in a given frame, the traffic-related objects (e.g. vehicles and pedestrians in this paper) are detected by deep-learning model, and we extract the scene as a series of frames from when the vehicle enters the scene to when it exits. When multiple vehicles are in motion, we treat each vehicle's presence in the scene as a separate scene. Pedestrians may or may not be present in each scene.

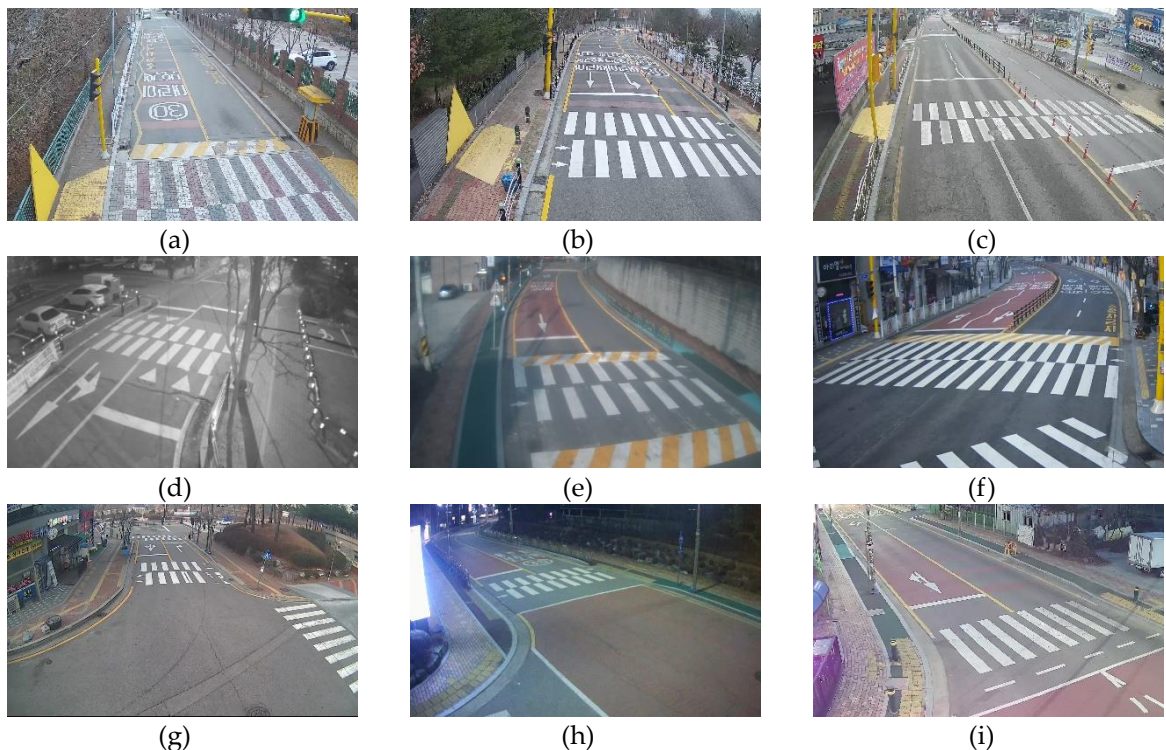

**Figure 2.** Actual CCTV views in (a) Spot A to (i) Spot I

To detect and segment these objects in the frames, we used a pre-trained mask R-CNN model, an extension of faster R-CNN. In our experiment, we used the Detectron 2 platform, as implemented by Facebook AI Research (FAIR) [18]. It is based on PyTorch deep learning framework, and has a much faster processing time than those of other existing platforms [19]. In addition, we used the pre-trained mask R-CNN model (ResNet-101-FPN) by Microsoft common objects in context (MS COCO) image dataset [20]. Our primary objective for using deep learning algorithms is to detect only traffic-related objects defined in this paper. Since the accuracy was close to perfect for these objects in our video footage, this pre-trained model did not need to be trained further for our purposes.

As a result, we extracted about 50,000 scenes from the entire video dataset, and used 45,890 scenes involving traffic-related objects as seen in Table 2. Each scene spanned approximately 38 frames, or 1.38 seconds. The majority of scenes captured only passing cars, while "interactive scenes" involved both vehicles and pedestrians in the scene at the same time.

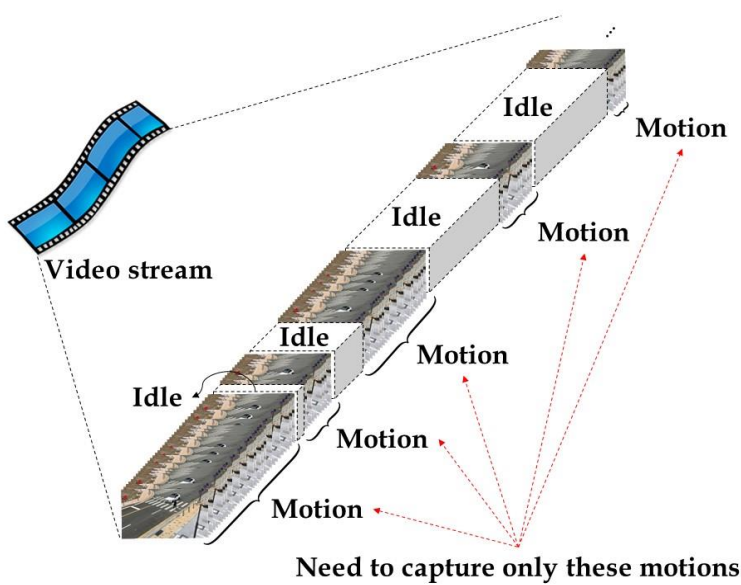

**Figure 3.** Composition of the actual video stream

**Table 2.** The number of the extracted scenes after preprocessing

| Spot code | # of scenes (After preprocessing) | | # of total frames | Avg. frames in one scene (ranges) |
|---|---|---|---|---|
| | Car-only scenes | Interactive scenes | | |
| A | 4,221 | | 136,189 | 32.26 frames (1.29 sec) |
| | 2,681 | 1,540 | | |
| B | 2,908 | | 86,249 | 29.66 frames (1.18 sec) |
| | 1,721 | 1,187 | | |
| C | 4,111 | | 382,980 | 93.16 frames (3.72 sec) |
| | 2,321 | 1,790 | | |

| | 6,955 | | 219,240 | 31.52 frames |
| --- | --- | --- | --- | --- |
| D | 4,633 | 2,322 | | (1.05 sec) |
| E | 3,876 | | 125,935 | 32.49 frames |
| | 2,481 | 1,395 | | (1.08 sec) |
| F | 7,587 | | 377,752 | 44.51 frames |
| | 6,494 | 1,093 | | (1.48 sec) |
| G | 5,612 | | 175,247 | 31.22 frames |
| | 3,533 | 2,079 | | (2.08 sec) |
| H | 2,845 | | 47,468 | 16.68 frames |
| | 1,843 | 1,002 | | (1.11 sec) |
| I | 7,775 | | 260,260 | 33.47 frames |
| | 4,572 | 3,203 | | (1.34 sec) |

## 2.2. Preprocessing

In this section, we describe the preprocessing module which consists of three steps: 1) contact points recognition; 2) perspective transform; and 3) object tracking and indexing.

Typically, road-deployed CCTV cameras record from oblique views, so it is difficult to precisely extract their behavioral features such as speeds and positions. To solve this, we transform the frame perspectives into an overhead view, using recognized contact points at the front center of the vehicles and between the feet of pedestrians. The more detailed procedures for this transformation are explained in our previous studies [14], [21]. In this paper, we focus on object tracking and indexing process.

The tracking algorithm from our previous work used the threshold and minimum distance methods [21]. This algorithm has two shortcomings, as seen in Figure 4. It only accounts for distance when postulating the location which an object can move to in the next frame, prioritizing the closest object rather than the most likely one. Such errors can cascade, with other objects regarded as disappearing out of frame, if their distance to the remaining positions is greater than the threshold.

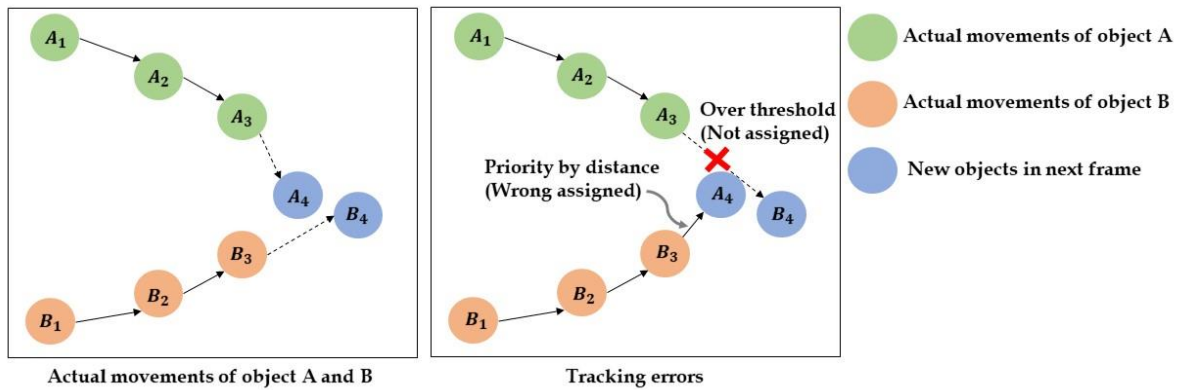

**Figure 4.** Example of the actual movements of two objects (left); and tracking errors (right)

To reduce these errors, we applied a modified Kalman filter method to more accurately track objects from frame to frame. Many research have been conducted on object tracking and indexing in various fields of computer science and transportation [22]–[24]. In particular, Kalman filters have been used in a wide range of engineering applications such as computer vision and robotics. They can efficiently calculate the state estimation process [25] and can be applied to estimate the unknown current or future states of the objects in video [26]. A Kalman filter calculates the next position of object by repeatedly performing two steps: 1) state prediction; and 2) measurement update. In the state prediction step, the current object's parameter values are predicted using previous values such as positions and speeds. In the measurement update step, the parameter values of the current object are updated by using the prior predicted values and information obtained about the current object position.

The tracking and indexing algorithm used in this study consists of two parts: 1) estimating the candidate points based on smoothing; and 2) assigning object in the next frame by calculating and comparing distances. First, we smooth the existing trajectory points using a Kalman filter to make positions and speeds more consistent. Then, we predict the next location of the trajectory, and calculate all distances between this and the candidate locations in the next frame, choosing the closest match.

For example, Figure 4 illustrates the actual movements of two objects, A and B, in multiple consecutive frames. Assume that we have already connected the trajectories of A and B from frames 1-3 so far, and are trying to correctly assign $A_4$ and $B_4$ to their correct trajectories. As represented in Figure 5, we smoothed the trajectories through frames 1-3, and predict the object's position in frame 4. The estimated points are denoted with apostrophe such as $A'_1$ $A'_2$, and $B'_3$ and the estimated target objects are denoted C, D, E, and F. Next, we calculate the distances between the origin target objects and the estimated target objects, denoted $Dist(origin\ target\ object, estimated\ target\ object)$. Finally, the target object with smallest distance from its prediction is assigned to the trajectory, and this process is repeated until the last frame in scene.

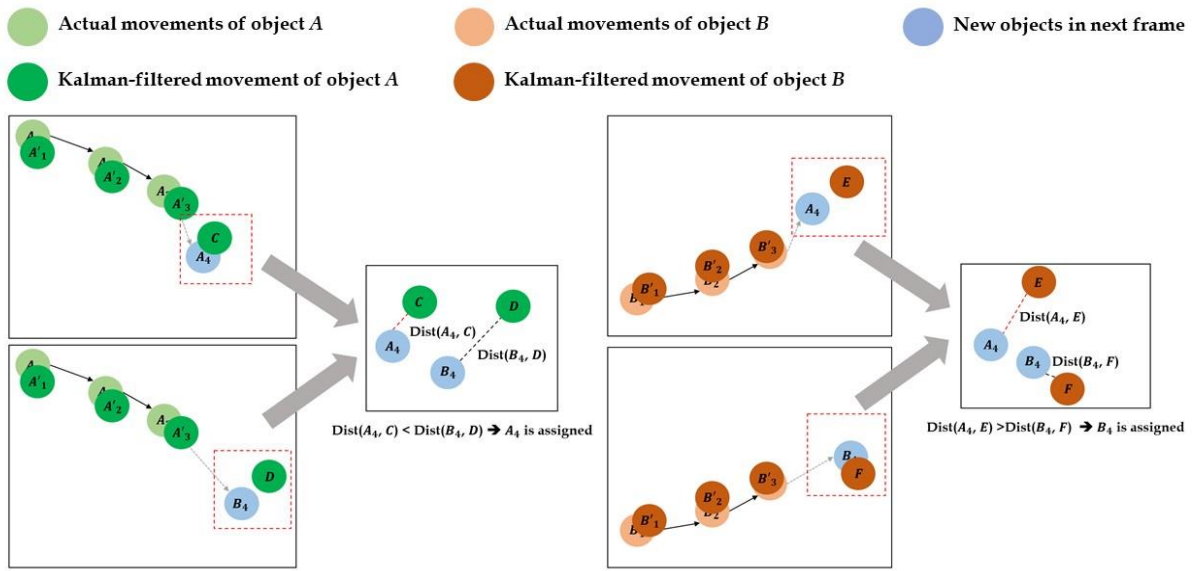

**Figure 5**. Process of object tracking and indexing algorithm for object A (left) and object B (right)

## 2.3. Automated Object Behavioral Feature Extraction

In this section, we describe which behavioral features were extracted and how to automate these processes. In general, there are two approaches to extracting features from video: 1) momentary feature; and 2) overall scene feature. A momentary feature is extracted at the frame-level, such as instantaneous speed; an overall scene feature is extracted at the scene-level, and describes the overall trajectory of the objects in the scene, such as the average vehicle speed, or the sequence of speeds measured at different times of the scene. Although we analyzed momentary features in our previous studies ([14], [21]), the primary goal of this study is to clarify the potential traffic risks presented by the behavior of drivers, so we extracted features at the scene-level.

Meanwhile, features can also be divided into single object features and interactive features. Single object features are characteristics of a single object without relation to other objects, such as speeds and positions; interactive features are measured between two objects, usually a vehicle and a pedestrian, such as the distance between them and pedestrian safety margin (PSM). In our experiment, we extracted ten features that could relate with potential traffic risk as seen in Table 3, and the extracting methods are described below in details.

Vehicle and pedestrian speeds: In general, object speed is a basic measurement that can signal potential risky situations. Car speed is a significant risk factor for pedestrian fatalities, and has a close relationship with crash severity in car-to-pedestrian collisions [27], [28]. Speed limits in our all testbeds were 30 km/h. A large number of detected vehicles traveling over the limit at any point, especially in school zones, contributes to high potential risk at that location. Meanwhile, pedestrian speed alone is not a direct indicator of such risks, but we may find important correlations and interactions with other features such as car speed and vehicle-pedestrian distance.

**Table 3.** The extracted features in our experiment

| Target object | Feature name | Description | Example |
|---|---|---|---|
| Vehicle | **Speed list** | ● Vehicle speeds change by frames<br>● Unit: km/h | ● [14.3, 12.0, 9.8, 4.3, 7.8, 12.1…] |
| | **Position list** | ● Vehicle positions change based on crosswalk by frames<br>● Represented as "before crosswalk", "on crosswalk" or "after crosswalk" | ● [before crosswalk, on crosswalk]<br>● [before crosswalk, on crosswalk, after crosswalk] |
| | **Acceleration list** | ● Vehicle accelerations change by frames<br>● Represented as "acceleration (acc)", "deceleration (dec)" or "no change (nc)" | ● [acc, nc]<br>● [nc]<br>● [acc, nc, acc] |
| | **Crosswalk distance list** | ● Distance changes between vehicles and crosswalks by frame<br>● Unit: m | ● [4.1, 3.3, 1.9, …] |
| | **Car stop before crosswalk** | ● Whether the vehicles stopped before passing the crosswalk in one scene<br>● Represented as "stop" or "no stop" | ● stop<br>● no stop |
| Pedestrian | **Speed list** | ● Pedestrian speeds change by frame<br>● Unit: km/h | ● [2.3, 2.0, 1.9, …] |
| | **Position list** | ● Pedestrian positions change by frames<br>● Represented as "sidewalk", "crosswalk" or "CIA (crosswalk influenced area)" | ● [sidewalk, CIA, sidewalk]<br>● [crosswalk] |
| **Vehicle-pedestrian interaction** | **Distance list** | • Distance changes between vehicle and pedestrian by frame<br>• Unit: m | • [4.1, 3.3, 1.9, …] |
| | **Relative position list** | • Relative positions list between vehicle and pedestrian by frame<br>• "Front" means pedestrian is in front side of the car, and "Behind" means pedestrian is back side of the car | • [Front, Front, Front, Behind, Behind]<br>• [Behind, Behind, Front] |
| | **Pedestrian safety margin** | • Pedestrian safety margin in one scene<br>• Unit: sec. | • 3.2<br>• -1.5 |

Object speed can be obtained from an assembled trajectory by dividing distance between its position in two consecutive frames by the time interval. In this case, the pixel distance between point $i$ in $j^{th}$ and $(j+1)^{th}$ frames in x-y plane, $D_{pixel}(point_i^j, point_i^{(j+1)})$, is computed by Euclidean distance method, and converted into real-world distance units such as meters. We infer pixel-per-meter constant, denoted as P, by dividing the pixel length of the crosswalk ($l_{pixel}$) by the actual length of it ($l_{world}$); we measured the actual lengths of crosswalks in field visits. For example, if the length of crosswalk is 15 m, and pixel length is 960 pixels, 1 meter is about 46 pixels (= 960 / 15).

Meanwhile, the frame intervals between trajectory points must be converted to real-world seconds. The time conversion constant (F) is computed by dividing the skipped frames by FPS. For example, if the video is recorded at 11 FPS, and we sampled every 5th frame, the time interval F is equal to 5/11. Finally, $i^{th}$ object's speed in $j^{th}$ and $(j+1)^{th}$ frames can be calculated as follows:

$$Speed_i^{j,(j+1)} = \frac{D_{pixel}(point_i^j, point_i^{(j+1)})}{F * P} \quad (m/s) \tag{3}$$

Finally, we convert these measurements into km/h, and apply them to all frames in the scene to obtain the instantaneous object speeds in each frame. As a result, the speed list of object i in scene k consisting of j frames is represented as:

$$velList_{k,i} = [\ speed_i^{1,2}, speed_i^{2,3}, speed_i^{3,4} \dots speed_i^{(j-1),j}] \tag{4}$$

<u>Vehicle and pedestrian positions</u>: The object positions are also important to investigate the potential traffic risks. A pedestrian on the road, even when cars are moving at slow speed, may be more at risk than a pedestrian on the sidewalk when cars are moving at high speed. In this study, car positions are categorized into three areas: "before crosswalk", "on crosswalk", and "after crosswalk", and pedestrian positions are categorized into four areas using their coordinates: "sidewalk", "crosswalk", "crosswalk influenced area (CIA)", and "road". CIA refers to the road area adjacent to the crosswalk, where pedestrians often enter while crossing the road [29]–[31]. In this study, we encompassed CIA with a buffer of ~3 meters on either side of the crosswalk.

<u>Vehicle acceleration list</u>: Vehicle accelerations and their changes during the scene are important factors to consider; if many vehicles maintain their speed or accelerate while approaching the crosswalk, this increases the risk to pedestrians. Ideally, we would expect to see cars decelerate near crosswalks, especially when pedestrians are present. In our experiment, we categorized vehicle accelerations as "acc", "dec", and "nc" by considering only speed changes. First, we smooth the speed sequence (see Figure 6) using a low-pass filter method, commonly used to reduce the rapid fluctuation of the signal that may result from the imprecision of object positioning from the image processing algorithm [32], [33]. This results in the filtered speed list, $F(velList_{k,i})$, with the filtered values, $f(velList_i^{j,(j+1)})$, where the subscripts $k$ and $i$ are scene number and object number in this scene, respectively.

Next, we calculated slope changes in the graph (means vehicle acceleration in time- speed graph) from when the vehicle enters the scene to when it reaches the crosswalk. We classified these as a sequence of acceleration states, with positive slopes yielding "acceleration", negative as "deceleration", and close to zero as "no change". This procedure can be written in mathematic equations as follows:

$$Acc_i^j = \begin{cases} \text{"acc"}, & f\left(vel_i^{(j+1),(j+2)}\right) - f\left(vel_i^{j,(j+1)}\right) > \varepsilon \\ \text{"dec"}, & f\left(vel_i^{(j+1),(j+2)}\right) - f\left(vel_i^{j,(j+1)}\right) < \varepsilon \\ \text{"nc"}, & otherwise \end{cases} \quad (5)$$

<u>Vehicle stop before crosswalk</u>: This feature indicates whether the vehicles came to a stop, before passing the crosswalk. Vehicles at these locations were required to stop once before passing the crosswalk, with or without pedestrians present. In practice, since the values of the extracted speeds have noise, we used a concept of "speed tolerance" to detect stops. The details on speed tolerance are described in our previous study, [14].

<u>Crosswalk distance and vehicle-pedestrian distance lists</u>: Crosswalk distance list means the distance changes between vehicles and crosswalk by frame, while vehicle-pedestrian distance list measures the sequence of distances between the vehicle and nearest-pedestrian by frame. Distances between vehicle i and pedestrian p are ordered by frame as follows:

$$dist_{i,p}^j = \frac{D_{pixel}\left(vehicle_i^j, pedestrian_p^j\right)}{P} \, (m) \quad (6)$$

$$distChng_{k,i,p} = [\, dist_{i,p}^1, \, dist_{i,p}^2, \, dist_{i,p}^3, \, ..., dist_{i,p}^j \,] \quad (7)$$

where the subscripts k and j are scene number and the frame order, respectively.

These distance-sequences alone are not factors for potential risk, but when compared with other features, we may identify dangerous situations. For example, Figure 7 (a) and (b) show two scenes as vehicle speeds plotted against vehicle-pedestrian distances while the pedestrian was on the crosswalk. In these examples, assume that the vehicle speed is not considered if it does not exceed the speed limit, and only investigate its changes by vehicle-pedestrian distance.

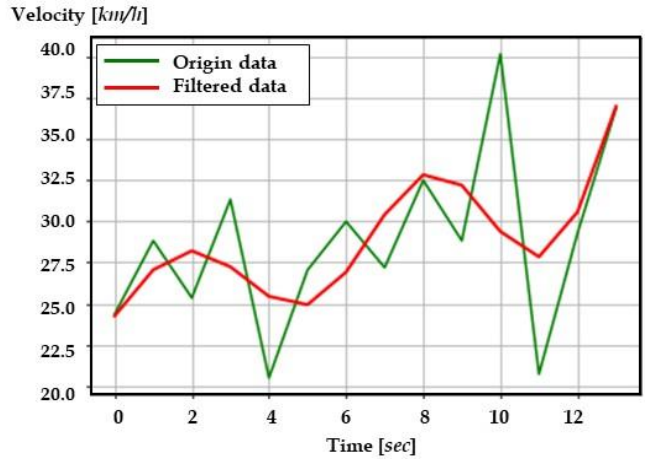

**Figure 6.** The origin speeds (green line) and the filtered data (red line)

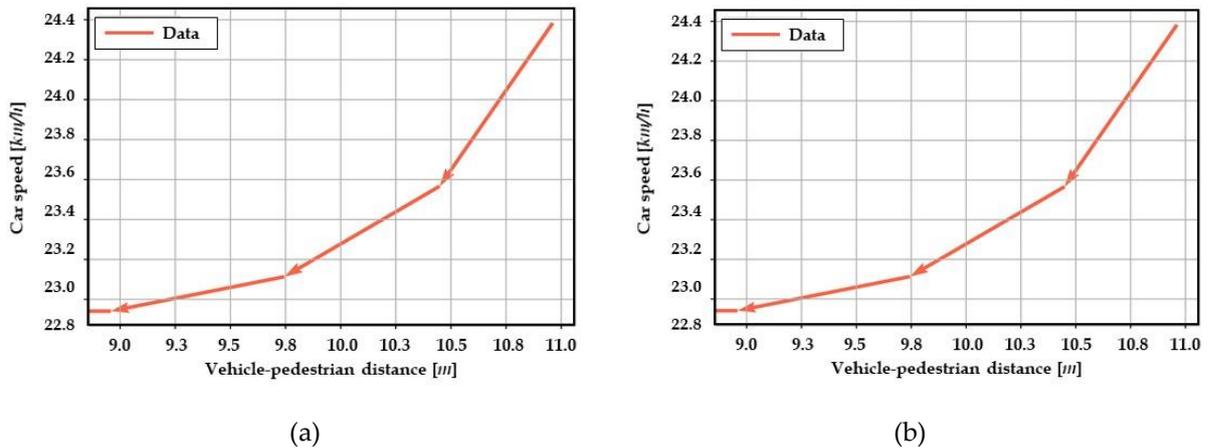

|(a)|(b)|

**Figure 7.** Examples of analyzing vehicle-pedestrian distance and other features

In Figure 7 (a), we can observe that as the vehicle approached the pedestrian, its speed decreased rapidly, then accelerated again immediately after the pedestrian passed. Although the vehicle slowed down when needed, it also accelerated rather rapidly even before the pedestrian had safely reached the sidewalk. In Figure 7 (b), the vehicle slows down as it approaches the pedestrian, and the speed is under the speed limit (almost 30 km/h). Now, we cannot determine which is more dangerous, but when considering only patterns of vehicle speeds, Figure 7 (a) is a pattern of re-acceleration after deceleration, and Figure 7 (b) is a pattern of continuous deceleration. These figures are just examples that have possibility to identify dangerous situations using the shapes of this features with others together.

<u>Relative position change between vehicles and pedestrians</u>: This describes the positional relationship between vehicles and pedestrians. If a pedestrian is in front of the car, they are at greater risk than if they were behind the car. We determine the relative positions between them by comparing their contact points, along with the position and direction of the vehicles.

This alone is not an obvious signal for risk, but when analyzed together with other features such as vehicle speed and pedestrian position, we may find important correlations and interactions between them. For example, a pedestrian who is behind a vehicle and on the sidewalk is in a relatively safe position.

Pedestrian safety margin (PSM): There are various ways to define the concepts of PSM[34]–[37]. In this study, we defined PSM as the time difference from when a pedestrian crossed the conflict point and when the next vehicle arrived at the same conflict point [34], [38], [39]. Suppose a pedestrian reaches a conflict point at time $T_1$, and the vehicle arrives at the same conflict point at time $T_2$, then the PSM is $T_2 - T_1$. Smaller PSM values mean there is less margin for error to avoid a collision at the conflict point.

Since the goal of this study is to extract these behavioral features automatically, it is important to infer the conflict point as seen in Figure 8. In this study, we applied virtual lines connecting the same objects between consecutive frames, and used intermediate value theorem (IVT).

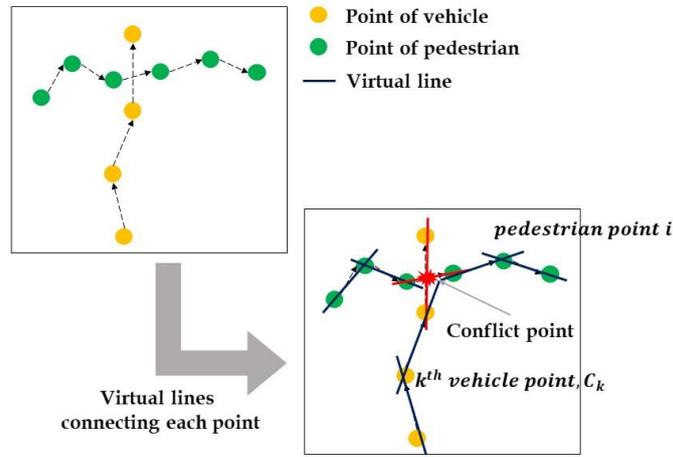

**Figure 8.** Expected conflict point in object trajectories

As represented as Figure 9, the process of PSM value extraction follows three steps: 1) drawing the virtual lines connecting the points of pedestrian in $i^{th}$ and $(i+1)^{th}$ frames, functionalized as linear function $f_{i,(i+1)}(x)$; 2) multiplying function values, $f_{i,(i+1)}(C_k)$ and $f_{i,(i+1)}(C_{k+1})$ where $C_k$ and $C_{k+1}$ are vehicles points, respectively; and 3) iterating steps 1 and 2 for all points in trajectories until $f_{i,(i+1)}(C_k) \times f_{i,(i+1)}(C_{k+1})$ is negative.

Applying IVT this way results in either a positive or negative value; if the result is positive, these points $i$ and $k$ are not in conflict. If it is negative, there is a conflict point between these points, and we can obtain the PSM values by calculating the difference between $i$ and $k$, and adjusting time unit from frames into seconds, as follows:

$$find \quad i, k \quad s.\,t. \quad f_{i,(i+1)}(C_k) \times f_{i,(i+1)}(C_{k+1}) < 0 \qquad (8)$$

$$PSM = \frac{(i-k)}{F} \ (sec) \qquad (9)$$

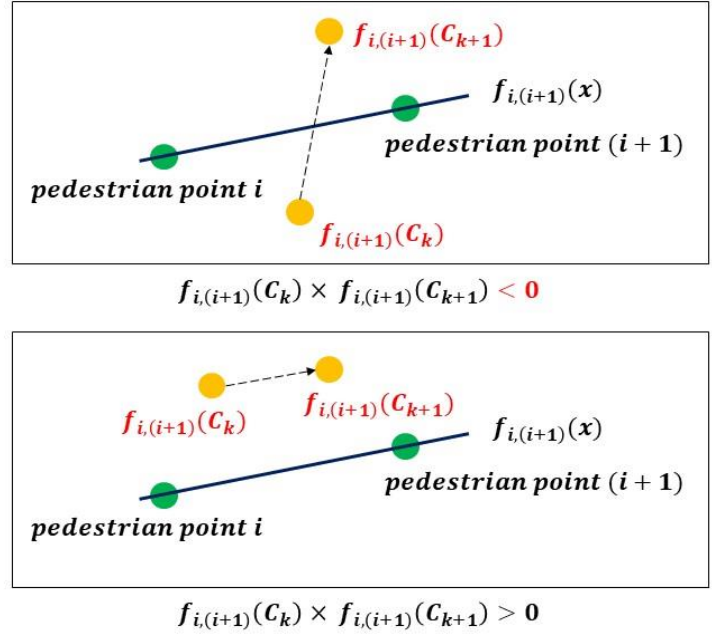

**Figure 9.** Process of finding conflict point by using IVT

## 3. Experiments and Results

With the proposed system, each objects' behavioral features can be extracted automatically. We applied this system in multiple crosswalks in Osan city, Korea. Then, we conducted statistical analysis on the extracted features, and behavioral characteristics of objects among these spots.

### 3.1. Experimental Design

In this section, we briefly describe how to validate results of our tracking and indexing algorithm, and our experimental design for analysis. First, in order to validate this algorithm, we defined success criteria, and manually counted all scenes with trajectories of objects that violated these criteria. Figure 10 (a) shows trajectories for correctly tracked objects.

As seen in these figures, the trajectories of objects should be continuous, and two or more objects should not cross each other. In addition, since this algorithm applied threshold method, if there are unallocated objects within the threshold range, they could be traced incorrectly.

Thus, we defined three criteria as follows:

- Connectivity: Are all of the objects connected in consecutive frames without breaks?

- Crossing: Are two or more objects, moving in parallel, traced separately without intertwining?

- Directivity: Do the objects follow their own paths without invading others' trajectories? This phenomenon may occur more frequently when adjusting threshold.

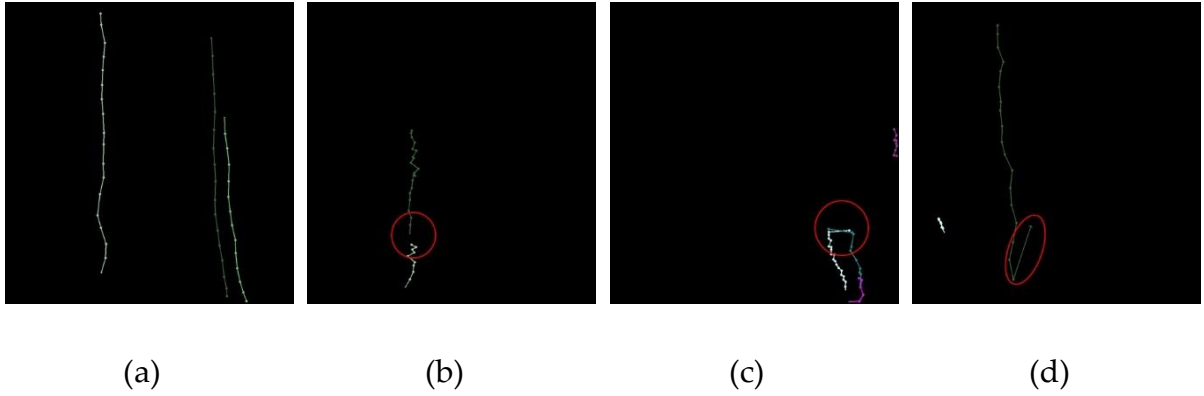

|      (a)       |       (b)        |       (c)        |       (d)       |

**Figure 10.** Trajectories of (a) the correctly tracked objects in scenes, and violating three criteria; (b) connectivity; (c) crossing; and (d) directivity

Figure 10 (b), (c), and (d) represent scenes that violate the above three criteria, respectively. The result of validation is shown in Table 4. We compared our tracking and indexing algorithm with our prior simple algorithm (see [21]). As a result, the average accuracy is about 3 percent higher than that of the existing method. In particular, by using the Kalman filter, the accuracy of directivity increased about 2 percent.

**Table 4.** Results of trajectory validation based on three criteria

| Spot code | # of scenes | Result of trajectory without Kalman filter (Car threshold = 100, Pedestrian threshold = 50) | | | | Result of Kalman filter-based trajectory (Car threshold = 60, Pedestrian threshold = 20) | | | |
|---|---|---|---|---|---|---|---|---|---|
| | | The number of error frames | | | Acc | The number of error frames | | | Acc |
| | | Connectivity | Crossing | Directivity | | Connectivity | Crossing | Directivity | |
| A | 4,789 | 45 | 98 | 305 | 0.91 | 25 | 66 | 194 | 0.94 |
| B | 3,195 | 35 | 75 | 285 | 0.88 | 21 | 58 | 201 | 0.91 |
| C | 5,311 | 32 | 112 | 401 | 0.90 | 22 | 74 | 298 | 0.93 |
| D | 7,304 | 49 | 155 | 491 | 0.90 | 40 | 101 | 347 | 0.93 |
| E | 4,261 | 54 | 98 | 358 | 0.88 | 41 | 59 | 256 | 0.91 |
| F | 8,036 | 61 | 187 | 652 | 0.89 | 45 | 111 | 515 | 0.92 |
| G | 6,259 | 55 | 138 | 499 | 0.89 | 35 | 77 | 398 | 0.92 |
| H | 3,295 | 25 | 59 | 441 | 0.84 | 14 | 32 | 387 | 0.86 |
| I | 7,940 | 35 | 90 | 595 | 0.91 | 28 | 47 | 457 | 0.93 |
| **Sum / Avg.** | | **291** | **1,012** | **4,027** | **0.89** | **271** | **635** | **3,053** | **0.92** |

Next, in our experiments, we analyzed the pedestrian's potential risks with three scenarios: 1) using distributions of car speed sand PSMs by spots; 2) investigating driver stopping behaviors when there are pedestrians on the crosswalk; and 3) considering PSMs together with stopping behaviors.

## 3.2. Results

3.2.1. Analyzing Car Speeds and PSM by Spots

Table 5 shows statistical values of average car speeds in each spot.

**Table 5.** Average vehicle speed information in all spots by scene types

| Spot code | All scenes | | | Types of scenes | |
|---|---|---|---|---|---|
| | Max. (km/h) | Min. (km/h) | Mean (km/h) | Car-only scene (Mean) | Interactive scene (Mean) |
| A | 71.3 | 3.6 | 18.2 | 20.5 km/h | 12.2 km/h |
| B | 87.5 | 4.4 | 24.5 | 25.9 km/h | 16.2 km/h |
| C | 75.4 | 6.5 | 36.5 | 41.7 km/h | 21.7 km/h |
| D | 79.7 | 4.1 | 18.1 | 18.4 km/h | 14.6 km/h |
| E | 68.1 | 2.2 | 22.3 | 22.3 km/h | 17.6 km/h |
| F | 51.3 | 3.9 | 20.9 | 21.2 km/h | 11.3 km/h |
| G | 63.9 | 9.4 | 14.0 | 14.2 km/h | 9.4 km/h |
| H | 59.2 | 3.3 | 21.4 | 21.5 km/h | 14.7 km/h |
| I | 70.2 | 7.4 | 33.8 | 34.8 km/h | 19.8 km/h |

The maximum average speeds are in the range of about 51.3 to 87.5 km/h, and minimum values range from 2.2 km/h to 9.4 km/h. The overall distributions are skewed right since many cars move slowly in these areas. The speed limit for all spots with school zones is 30 km/h. When considering mean values in all spots are near or under the regulation speed, these are reasonable values.

In general, cars tend to move faster when there are no pedestrians present, and slow down when there are pedestrians. We can observe these tendencies by separating the average vehicles speeds into car-only scenes and interactive scenes as seen in Table 5. In all spots, the speeds in interactive scenes are lower than those in car-only scenes.

Spot C is the only location where the average speeds exceeded the speed limit (30 km/h). This may be related with the number of lanes and whether a speed camera is deployed. First, Spot C has 4 lanes, more lanes than any other spot except Spot F; generally, higher speed limits apply when there are more lanes, but the speed limit in Spot C remains 30 km/h because it is designated as a school zone. Second, Spot F matches Spot C in the number of lanes, speed limit, signalized crosswalk, and school zone designation, but Spot F has a speed camera, missing from Spot C (refer Table 1). From this example, we can hypothesize that when the number of lanes increase, vehicle speeds increase, but a speed camera can suppress such tendency.

Next, we analyzed the extracted PSM distributions. Note that PSM counts how many seconds it takes for a car to pass through the same point after a pedestrian passes it, thus quantifying the potential risk of vehicle-pedestrian collision situation. In our experiment, we filtered out the negative values and only look at cars passing behind the pedestrians (Negative PSM values mean that the car passed before the pedestrian). Then, we differentiated between the signalized crosswalks (spots A, B, C, and F) vs unsignalized crosswalks (spots D, E, G, H, and I).

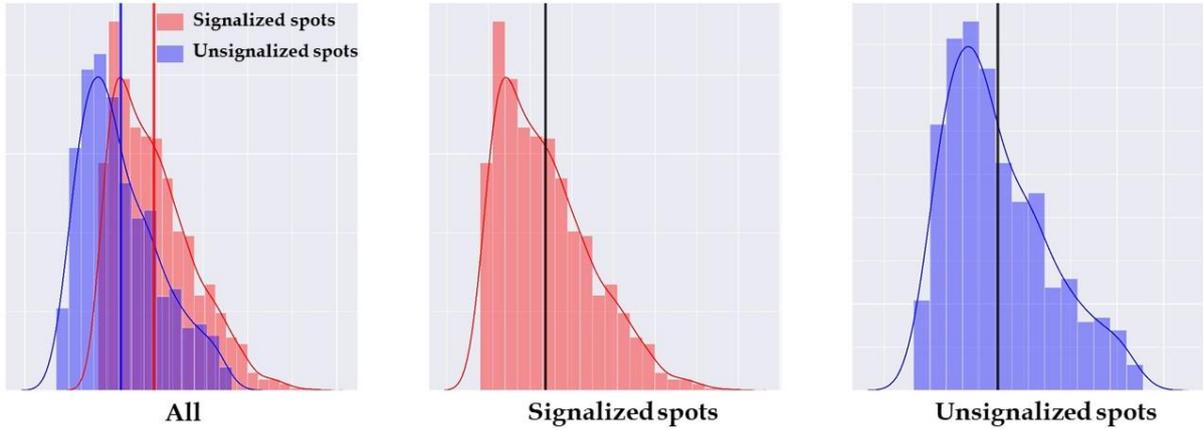

**Figure 11.** Distributions of PSMs in signalized and unsignalized spots

Figure 11 shows the distributions of positive PSM in all signalized vs unsignalized spots. It represents the ranges and mean values of PSM; PSMs were higher on average in signalized crosswalks than those in unsignalized crosswalks. In addition, peak of the distribution across all signalized spots is higher, since the traffic signal forces some time to pass before cars can cross the pedestrian's path. Without the signal, the distribution peaks closer to zero, indicating cars are not willing to wait and give pedestrian the safety margin before passing.

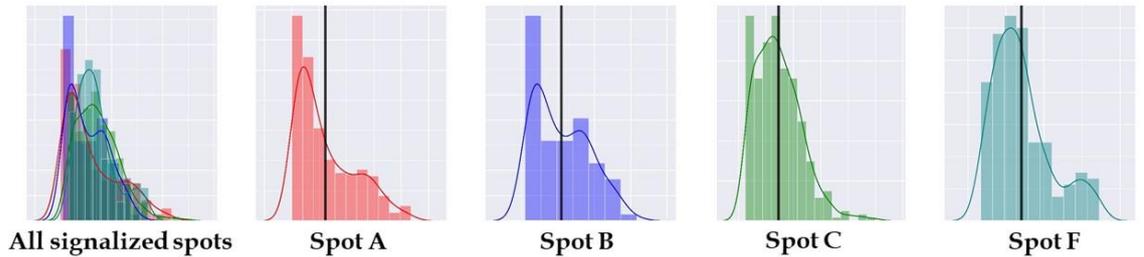

(a)

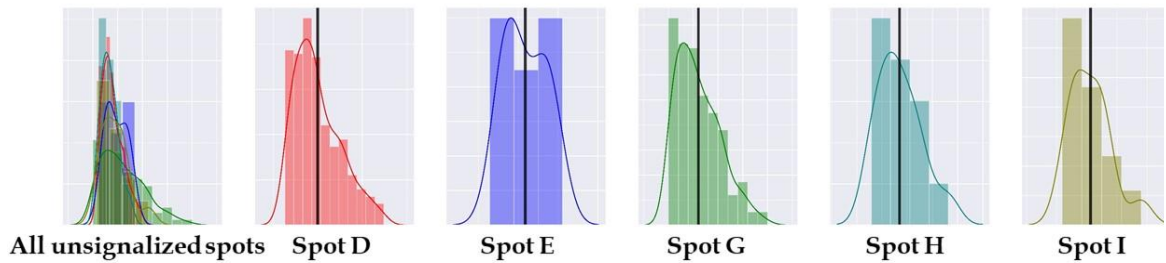

(b)

**Figure 12.** Distributions of PSMs in (a) signalized spots; and (b) unsignalized spots

Figure 12 (a) and (b) show distributions of PSM at each spot. In Figure 12 (a), we can observe that in signalized spots, wider roads lead to higher PSM, possibly because of longer signal cycles for pedestrian crossing. Spots C and D each have four lanes, wider than Spot A (2 lanes) and B (3 lanes), and their PSM distributions are further to the right.

Meanwhile, in unsignalized crosswalks, the overall distributions are similar to each other, and we did not observe a relationship between road width and PSM distribution. Spot G stood out with PSM distribution further right of the others; one reason could be its slower vehicle speeds overall. Since it is in a residential area, it has a particularly high floating population (especially students) during rush hour. In addition, there are road intersections close to either side of the crosswalk (see Figure 2 (g)), forcing slower speeds and more careful manuevering for vehicles, who in turn give pedestrians plenty of crossing time.

3.2.2. Analyzing Pedestrian's Potential Risk near Crosswalks based on Car Stopping Behaviors

In this section, we analyzed whether or not vehicles stopped before passing the crosswalk when a pedestrian was present, and the distance they stopped from the crosswalk. Generally, vehicles may stop for a variety of reasons such as parking on the shoulder, waiting for a traffic signal, or allowing pedestrians right-of-way. To precisely count the scenes when the driver stopped to ensure pedestrian safety, we chose 10 m as a baseline distance; if a car stopped within 10 m from the crosswalk, with a pedestrian in the crosswalk or CIA, we assumed they were reacting to the pedestrian's presence.

Figure 13 (a) and (b) show the percentages of vehicles that stopped within 10 m before passing the crosswalks when pedestrians crossed the streets in signalized and unsignalized spots, respectively. First, among signalized spots, Spot A has the lowest percentage of drivers stopping. The reason would be related with the width of lanes. Spot A has just two lanes, but other signalized spots have three or more lanes. It can be interpreted that the drivers in the narrow road are reluctant to wait for the signal, so they would violate the signal. Spot F has the highest percentages than those in other spots. It can be seen that the installation of the speed camera has a deterrent force that makes the drivers keep the signal well. In this experiment, we analyzed only behaviors of vehicles and pedestrians, not considering signal phases together. Note that the coexistence of the passing vehicle and crossing pedestrian implies that one of the traffic participants violates the traffic signal threatening driving safety regardless of the signal.

Meanwhile, in unsignalized spots, especially spots G and H, most drivers did stop before passing the crosswalk. Spot H had a relatively high stopping percentage, perhaps due to its safety features such as red urethane pavement and "school zone" lettering on the road, as well as safety fences on both sides of the road. Spot G also had a high stopping percentage. However, since there were no signal lights, drivers were less likely to perform the required safe behavior (stopping before the crosswalk until pedestrians have cleared the area). In particular, half or more of the drivers in spots D, E, and I failed to stop when pedestrians were on the road, despite their designation as school zones. In these spots, further proactive response seems necessary to encourage stopping for pedestrians, and prevent accidents before they occur.

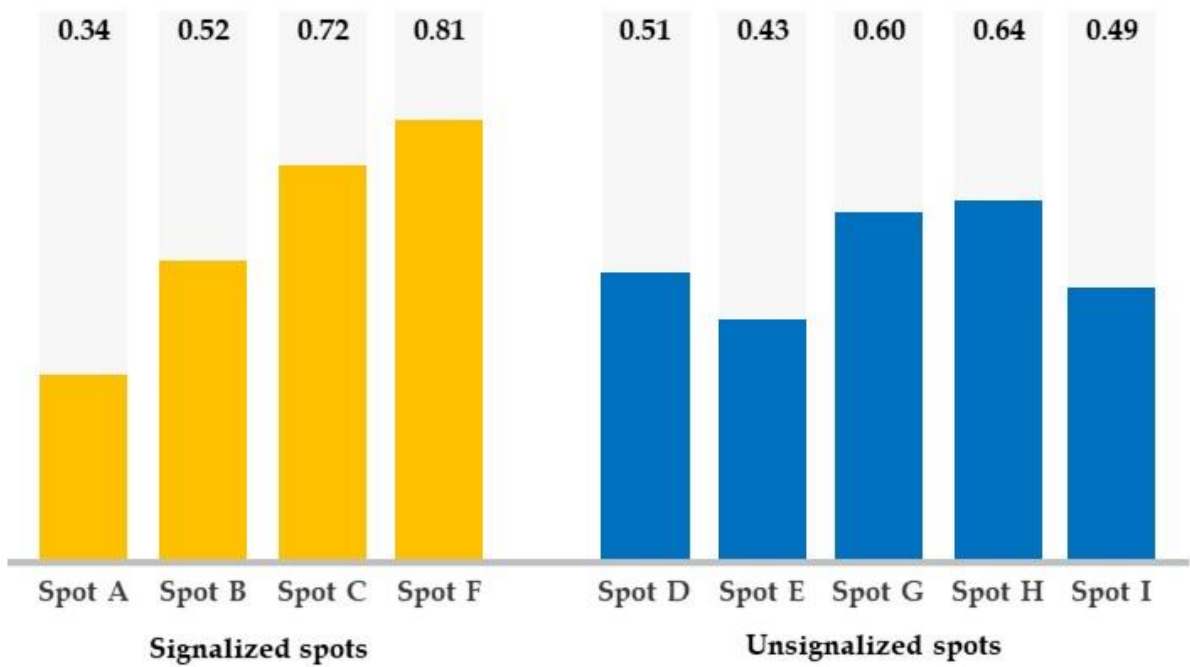

**Figure 13.** The percentages of drivers stopping within 10 m from crosswalks for scenes with pedestrians on crosswalks

3.2.3. Analyzing Car Behaviors with PSM and Car Stopping near the Unsignalized Crosswalk

In this section, we analyzed driver stopping behaviors with PSM values at unsignalized crosswalks. PSM is a simple feature that can provide implicative information for vehicle and pedestrian behaviors. Since PSM is the time difference from when a pedestrian passed a certain point and when the vehicle arrived at the same point, a positive PSM value means that the pedestrian crossed first, and a negative value means that the vehicle passed first. Since the latter implies that the vehicle failed to yield to the pedestrian in the crosswalk, negative PSM values generally present more risk than positive values. In either case, collision risk increases as PSM approaches zero. We only considered scenes in unsignalized spots in this section, since yielding behavior and PSM at signalized crosswalks greatly depend on the traffic signal at the time of encounter.

In our experiment, we studied scenes occurring within various ranges of PSM, and measured the likelihood of a vehicle stopping before the crosswalk with a pedestrian present (using 10 m as a baseline distance as in Section 3.B.2). First, we categorized the continuous PSM values into eight groups by signs and quartiles, using a combined distribution accounting for all scenes in the five unsignalized crosswalks.

However, simply merging these distributions would bias the result toward the distribution of higher-traffic areas. For example, if there were 100 and 800 scenes in two regions A and B, respectively, the merged distribution across these two regions would be more affected by scenes occurring in B. Thus, we calculated the weight of each distribution relative to the whole:

$$w_i = 1 - \frac{|D_i|}{|D|} \tag{10}$$

where $|D|$ is total number of scenes in unsignalized spots (spots D, E, G, H, and I) and $|D_i|$ is number of scenes in each spot. We then multiplied by $w_i$ to normalize the scene frequencies in spot $i$. As a result, Figure 14 (a) and (b) represent the combined, weighted distributions of PSM values across all scenes in unsignalized crosswalks.

From these distributions, we split between positive and negative PSM values, and within each by quartile, to yield the following PSM ranges: 1) under -4.92; 2) -4.92 to -3.04; 3) -3.04 to -2.03; 4) -2.03 to 0; 5) 0 to 1.25; 6) 1.25 to 2.29; 7) 2.29 to 3.91; and 8) over 3.91, denoted by ranges 1 to 8, respectively. Then, we compared the stopping percentages within each PSM range.

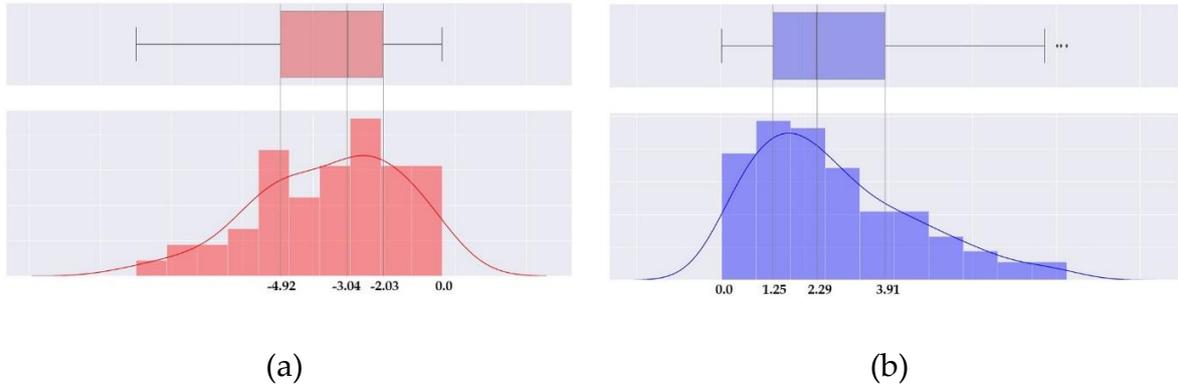

(a)                                    (b)

**Figure 14.** Distributions of PSMs in (a) signalized; and (b) unsignalized spots

In Figure 15, ranges 1, 2, 3, 6, 7, and 8 are relatively distant groups from zero, and ranges 4 and 5 present the greatest risk, with safety margins within 1-2 seconds. We can observe that as margins increase, vehicles are less likely to stop at the crosswalk.

Ideally, for small but positive PSM scenes, we would want to see the highest stopping percentages in order to minimize risk of collision with pedestrians. Yet, within range 5 (PSM between 0 and 1.25 sec), most cars in spot E did not stop. This could result from two possible behaviors: 1) drivers did not stop, but decelerated while passing ahead of pedestrians, or 2) drivers did not stop nor decelerate, and narrowly avoided collisions with pedestrians. Thus, Spot E represents an anomaly, since stopping percentages for other spots in these low-margin ranges are at least 50%; since it presents greater risk of collision, we would want to understand why and proactively address the issue.

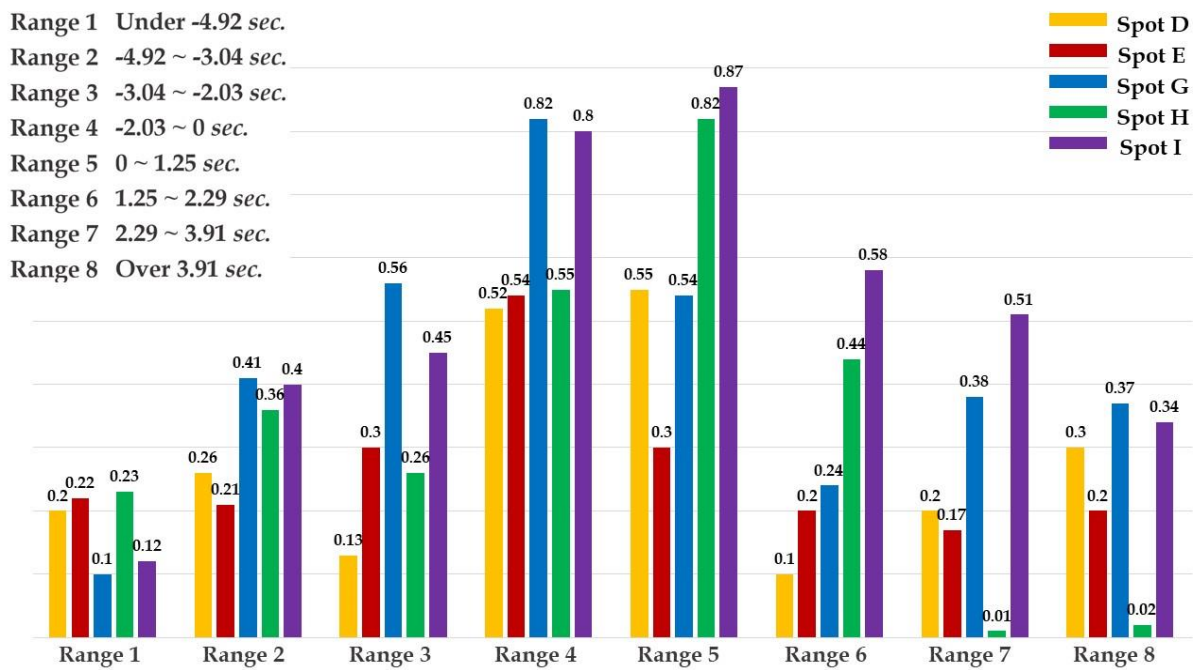

**Figure 15.** The percentages of drivers stopping before the crosswalk by PSM range

Meanwhile, we can see that at larger PSM margins, especially ranges 2, 3, 7, and 8, stopping percentages are highest in spots G and I. We hypothesize that this is because G and I have no fences separating the road from the sidewalk, unlike the other unsignalized spots. Without the fences, drivers may be forced to drive more cautiously through the area, since pedestrians could potentially enter the road at any point along the approach to the crosswalk. In these areas, adding safety features such as sidewalk fences could negatively affect the behavior of vehicles and pedestrians, by removing the uncertainty that forces driver caution and more frequent stopping.

### 3.3. Discussions

The proposed system in this research had three main objectives: (1) to automatically collect video clips (scenes) from video streams by using frame difference method on an end-point server with direct access to the image feed; (2) to automatically extract the objects' behavioral features using deep learning model and computer vision techniques; and (3) to analyze these behavioral features to identify pedestrian's potential risk and possible environmental and road design factors that affect risk. Unlike our previous study [21], this research analyzed risky situations at the scene level, and expanded the scale to more cameras over longer timeframes, and capturing diverse road environments, such as signalized and unsignalized crosswalks.

In our experiments, we extracted time- and distance-based sequences of various behavioral features affecting potential risk such as vehicle and pedestrian speeds, vehicle accelerations, and PSM. In order to observe how sensitive drivers were to the risk of pedestrian collision, we categorized scenes as car-only vs. vehicle-pedestrian interactive scenes. Then, we performed three analyses: (1) distributions of the average car speeds and PSMs by spots; (2) percentages of vehicles stopping when pedestrians are present in or near the crosswalk; and (3) stopping behaviors relative to PSM. We observed how vehicle speeds responded to road environments, and how they changed when approaching pedestrians.

One limitation of this system is the lack of an interface to perform a comprehensive analysis of various situations. For example, the size and complexity of the generated dataset make it difficult to answer questions such as: "at unsignalized crosswalks, when the average vehicle speed is between 30 to 40 km/h between 8 am to 9 am, and pedestrians are present, what is the acceleration state of the vehicles in each spot?" or "when PSM is in range of -1 to 0, what were vehicle speeds in school zones in the evening?". In order to address these challenges, we need to classify the given behavioral features according to their characteristics to enable multidimensional analysis, such as on-line analytical process (OLAP) and data mining techniques. This would allow administrators (e.g. transportation engineers or city planners) to interpret the behavior features, understand existing areas, design alternative roads/crosswalks/intersections, and test the impact of these physical changes.

## 4. Conclusions

In this study, we proposed a comprehensive potential risky scene analysis model for automatically extracting traffic objects' behavioral features from raw video footage. The research scope in this paper is from data sources to behavioral feature extraction modules, and multidimensional analysis will be covered in our future research. The core methodologies are 1) recognizing the objects' precise contact points even when video footage was recorded from oblique viewpoints; 2) extracting various behavioral features by time and distance based on object tracking and indexing algorithm; and 3) analyzing them at the scene level, to understand how drivers behave in each environment and situation. We validated the feasibility of the proposed analysis system by implementing it with Detectron 2 platform and OpenCV libraries, applying it to actual video data from crosswalks in Osan city, Korea, and visualizing and interpreting these scenes at each spot.

Our proposed analysis model elicits useful information that can be obtained without additional expenditure in new camera infrastructure, by extracting anonymized movement trajectories of vehicles near pedestrians. Moreover, these analyses can provide powerful and useful information for decision makers to improve the road environment safer. However, the system itself would not identify the best control or traffic calming measures to prevent traffic accidents. We hypothesize that it can provide practitioners with enough clues to support further investigation through other means. Furthermore, traffic engineers and urban designers must collaborate using these clues to improve the safety of the spaces. Our goal in developing this system was to aid in this collaboration, by making it faster, cheaper, and easier to collect objective information about the behavior of drivers at places where pedestrians face the greatest risks.

**Supplementary Materials**: Not Applicable

# References

1. G. T. S. Ho, Y. P. Tsang, C. H. Wu, W. H. Wong, and K. L. Choy, "A computer vision-based roadside occupation surveillance system for intelligent transport in smart cities," Sensors (Switzerland), vol. 19, no. 8, 2019, doi: 10.3390/s19081796.

2. M. D. Lytras and A. Visvizi, "Who uses smart city services and what to make of it: Toward interdisciplinary smart cities research," Sustain., vol. 10, no. 6, pp. 1–16, 2018, doi: 10.3390/su10061998.

3. F. Akhter, S. Khadivizand, H. R. Siddiquei, M. E. E. Alahi, and S. Mukhopadhyay, "Iot enabled intelligent sensor node for smart city: Pedestrian counting and ambient monitoring," Sensors (Switzerland), vol. 19, no. 15, 2019, doi: 10.3390/s19153374.

4. Y. Yang and M. Ning, "Study on the Risk Ratio of Pedestrians' Crossing at Unsignalized Crosswalk," CICTP 2015 - Efficient, Safe, and Green Multimodal Transportation - Proceedings of the 15th COTA International Conference of Transportation Professionals. pp. 2792–2803, 2015, doi: 10.1061/9780784479292.257.

5. T. Gandhi and M. M. Trivedi, "Pedestrian protection systems: Issues, survey, and challenges," IEEE Trans. Intell. Transp. Syst., vol. 8, no. 3, pp. 413–430, 2007, doi: 10.1109/TITS.2007.903444.

6. V. Gitelman, D. Balasha, R. Carmel, L. Hendel, and F. Pesahov, "Characterization of pedestrian accidents and an examination of infrastructure measures to improve pedestrian safety in Israel," Accid. Anal. Prev., vol. 44, no. 1, pp. 63–73, 2012, doi: 10.1016/j.aap.2010.11.017.

7. P. Olszewski, P. Szagała, M. Wolański, and A. Zielińska, "Pedestrian fatality risk in accidents at unsignalized zebra crosswalks in Poland," Accid. Anal. Prev., vol. 84, pp. 83–91, 2015, doi: 10.1016/j.aap.2015.08.008.

8. K. Haleem, P. Alluri, and A. Gan, "Analyzing pedestrian crash injury severity at signalized and non-signalized locations," Accident Analysis and Prevention, vol. 81. pp. 14–23, 2015, doi: 10.1016/j.aap.2015.04.025.

9. T. Fu, W. Hu, L. Miranda-Moreno, and N. Saunier, "Investigating secondary pedestrian-vehicle interactions at non-signalized intersections using vision-based trajectory data," Transp. Res. Part C Emerg. Technol., vol. 105, no. September 2018, pp. 222–240, 2019, doi: 10.1016/j.trc.2019.06.001.

10. T. Fu, L. Miranda-Moreno, and N. Saunier, "A novel framework to evaluate pedestrian safety at non-signalized locations," Accid. Anal. Prev., vol. 111, no. March 2017, pp. 23–33, 2018, doi: 10.1016/j.aap.2017.11.015.

11. R. Ke, J. Lutin, J. Spears, and Y. Wang, "A Cost-Effective Framework for Automated Vehicle-Pedestrian Near-Miss Detection Through Onboard Monocular Vision," IEEE Comput. Soc. Conf. Comput. Vis. Pattern Recognit. Work., vol. 2017-July, pp. 898–905, 2017, doi: 10.1109/CVPRW.2017.124.

12. B. Murphy, D. M. Levinson, and A. Owen, "Evaluating the Safety In Numbers effect for pedestrians at urban intersections," Accid. Anal. Prev., vol. 106, no. May, pp. 181–190, 2017, doi: 10.1016/j.aap.2017.06.004.

13. B. R. Kadali and P. Vedagiri, "Proactive pedestrian safety evaluation at unprotected mid-block crosswalk locations under mixed traffic conditions," Saf. Sci., vol. 89, pp. 94–105, 2016, doi: 10.1016/j.ssci.2016.05.014.

14. B. Noh, W. No, J. Lee, and D. Lee, "Vision-based potential pedestrian risk analysis on unsignalized crosswalk using data mining techniques," Appl. Sci., vol. 10, no. 3, 2020, doi: 10.3390/app10031057.

15. "NATIONAL LAW INFORMATION CENTER." http://www.law.go.kr/lsSc.do?tabMenuId=tab18&query=#J5:13]. (accessed May 05, 2020).

16. L. H, D. J, W. R, Z. H, and Zh. B, "Combining background substraction and three-frame difference to detect moving object from underwater video." OCEANS 2016-Shanghai, pp. 1–5, 2016.

17. S. S. Sengar and S. Mukhopadhyay, "Moving object detection based on frame difference and W4," Signal, Image Video Process., vol. 11, no. 7, pp. 1357–1364, 2017, doi: 10.1007/s11760-017-1093-8.

18. "Facebook AI Research." https://ai.facebook.com/ (accessed Jan. 17, 2020).

19. "Github." https://github.com/facebookresearch/detectron2.

20. "COCO Dataset." http://cocodataset.org/#home (accessed Sep. 03, 2019).

21. B. Noh, W. No, and D. Lee, "Vision-based overhead front point recognition of vehicles for traffic safety analysis," UbiComp/ISWC 2018 - Adjunct Proceedings of the 2018 ACM International Joint Conference on Pervasive and Ubiquitous Computing and Proceedings of the 2018 ACM International Symposium on Wearable Computers. pp. 1096–1102, 2018, doi: 10.1145/3267305.3274165.

22. P. C. Besse, B. Guillouet, J. M. Loubes, and F. Royer, "Review and Perspective for Distance-Based Clustering of Vehicle Trajectories," IEEE Transactions on Intelligent Transportation Systems, vol. 17, no. 11. pp. 3306–3317, 2016, doi: 10.1109/TITS.2016.2547641.

23. Y. Guan, S. Penghui, Z. Jie, L. Daxing, and W. Canwei, "A review of moving object trajectory clustering algorithms," Artificial Intelligence Review, vol. 47, no. 1. pp. 123–144, 2017, [Online]. Available: https://link.springer.com/content/pdf/10.1007%2Fs10462-016-9477-7.pdf.

24. S. Zuo, L. Jin, Y. Chung, and D. Park, "An index algorithm for tracking pigs in pigsty," Industrial Electronics and Engineering, vol. 1. pp. 797–804, 2014, doi: 10.2495/iciee140931.

25. B. Haroun, L. Q. Sheng, L. H. Shi, and B. Sebti, "Vision Based People Tracking System," vol. 13, no. 11, pp. 582–586, 2019.

26. X. Sun, H. Yao, and S. Zhang, "A refined particle filter method for contour tracking," Visual Communications and Image Processing 2010, vol. 7744. p. 77441M, 2010, doi: 10.1117/12.863450.

27. "Stocker (2015) pedestrian safety and the built environment.pdf." .

28. H. Jeppsson, M. Östling, and N. Lubbe, "Real life safety benefits of increasing brake deceleration in car-to-pedestrian accidents: Simulation of Vacuum Emergency Braking," Accident Analysis and Prevention, vol. 111. pp. 311–320, 2018, doi: 10.1016/j.aap.2017.12.001.

29. M. A. Figliozzi and C. Tipagornwong, "Pedestrian Crosswalk Law: A study of traffic and trajectory factors that affect non-compliance and stopping distance," Accid. Anal. Prev., vol. 96, pp. 169–179, 2016, doi: 10.1016/j.aap.2016.08.011.
25

30. T. Fu, "A novel apporach to investigate pedestrian safety in non-signalized crosswalk environmets and related treatments," no. March, 2019.

31. V. P. Sisiopiku and D. Akin, "Pedestrian behaviors at and perceptions towards various pedestrian facilities: An examination based on observation and survey data," Transp. Res. Part F Traffic Psychol. Behav., vol. 6, no. 4, pp. 249–274, 2003, doi: 10.1016/j.trf.2003.06.001.

32. J. Sinclair, P. John Taylor, and S. Jane Hobbs, "Digital filtering of three-dimensional lower extremity kinematics: An assessment," J. Hum. Kinet., vol. 39, no. 1, pp. 25–36, 2013, doi: 10.2478/hukin-2013-0065.

33. A. Widmann, E. Schröger, and B. Maess, "Digital filter design for electrophysiological data - a practical approach," J. Neurosci. Methods, vol. 250, pp. 34–46, 2015, doi: 10.1016/j.jneumeth.2014.08.002.

34. C. Avinash, S. Jiten, S. Arkatkar, J. Gaurang, and P. Manoranjan, "Evaluation of pedestrian safety margin at mid-block crosswalks in India," Saf. Sci., vol. 119, no. September 2018, pp. 188–198, 2019, doi: 10.1016/j.ssci.2018.12.009.

35. X. Chu and M. R. Baltes, "Pedestrian Mid-block Crossing Difficulty Final Report," p. 79, 2001.

36. J. A. Oxley, E. Ihsen, B. N. Fildes, J. L. Charlton, and R. H. Day, "Crossing roads safely: An experimental study of age differences in gap selection by pedestrians," Accid. Anal. Prev., vol. 37, no. 5, pp. 962–971, 2005, doi: 10.1016/j.aap.2005.04.017.

37. R. Lobjois and V. Cavallo, "Age-related differences in street-crossing decisions: The effects of vehicle speed and time constraints on gap selection in an estimation task," Accid. Anal. Prev., vol. 39, no. 5, pp. 934–943, 2007, doi: 10.1016/j.aap.2006.12.013.

38. R. Almodfer, S. Xiong, Z. Fang, X. Kong, and S. Zheng, "Quantitative analysis of lane-based pedestrian-vehicle conflict at a non-signalized marked crosswalk," Transp. Res. Part F Traffic Psychol. Behav., vol. 42, pp. 468–478, 2016, doi: 10.1016/j.trf.2015.07.004.

39. J. D. Bullough and N. P. Skinner, "Pedestrian Safety Margins Under Different Types of Headlamp Illumination." pp. 1–14, 2009.